# Fast Iterative Five point Relative Pose Estimation

Johan Hedborg  Michael Felsberg
Computer Vision Laboratory
Linköping University, Sweden
https://www.cvl.isy.liu.se/

## Abstract

*Robust estimation of the relative pose between two cameras is a fundamental part of Structure and Motion methods. For calibrated cameras, the five point method together with a robust estimator such as RANSAC gives the best result in most cases. The current state-of-the-art method for solving the relative pose problem from five points is due to Nistér [9], because it is faster than other methods and in the RANSAC scheme one can improve precision by increasing the number of iterations.*

*In this paper, we propose a new iterative method, which is based on Powell's Dog Leg algorithm. The new method has the same precision and is approximately twice as fast as Nistér's algorithm. The proposed method is easily extended to more than five points while retaining a efficient error metrics. This makes it also very suitable as an refinement step.*

*The proposed algorithm is systematically evaluated on three types of datasets with known ground truth.*

## 1. Introduction

The estimation of the relative position and orientation of two cameras based on the image content exclusively is the basis for a number of applications, ranging from augmented reality to navigation. In many cases, it is possible to estimate the intrinsic parameters of the camera prior to the pose estimation. This reduces the number of unknowns to five. This particular case is the problem considered in this paper and the current state-of-the-art solution is the five point solver presented by Nistér [9] combined with a preemptive scoring scheme, [10], in a RANSAC loop, [3].

We propose to solve the five point problem iteratively using Powell's Dog Leg (DL) method, [11]. We show that a non-linear DL solver with five parameters is faster than the currently fastest five point solver [9]. The DL optimization method is a commonly used method for solving general non-linear systems of equations and least square problems.

Our formulation of the five point problem is based on the epipolar distance, [5], i.e. the distance of the points to the respective epipolar lines. By minimizing an error based on the epipolar distance instead of an error in 3D space, no structure estimation is required, thus leading to fewer unknown parameters and to a significant reduction in calculations.

### 1.1. Related work

In the case of calibrated cameras, the pose estimation problem is most accurately solved using the minimal case of five points within a robust estimator (RANSAC) framework, [9, 17, 12]. The five point problem is a well understood topic and there exists a wide variety of solutions. Among the first methods is the early work by [2] and more recent work can be found in [7, 18].

As a consequence of using the five point solver within a RANSAC loop, the speed of the solver is of great practical relevance because the accuracy can be improved by using more iterations, [5]. Therefore, the current state-of-the-art method is due to Nistér [9], as his method is capable of calculating the relative pose within a RANSAC loop in video real-time. However, to achieve this level of performance, a considerable implementation effort is required, [7].

Stewénius [18] improved the numerical stability of [9] by using Gröbner bases to solve a series of third order polynomials instead of solving a tenth order polynomial. However, this gain in stability comes with a performance penalty due to a 10x10 eigenvector decomposition.

An alternative to algebraic solvers are non-linear optimization methods, which find the solution iteratively, e.g. using the Gauss-Newton method, [14]. However, the Gauss-Newton method becomes unstable if the Hessian is ill-conditioned. To compensate for that the authors propose to include more than five points in the estimation of the pose, [14], section 4.2. The speed reported in [14] is prohibitively slower (about four orders of magnitude) than Nistér's implementation.

In terms of evaluation, many authors have been using synthetic data in order to have access to ground truth information, [9, 18, 12]. Using additive Gaussian noise in these synthetic datasets does not necessarily lead to realistic eval-uation results.

### 1.2. Main contributions

The main contributions of this paper are:

- We propose a new iterative method for solving the five point relative pose problem using the Dog Leg algorithm.
- The new method is nearly a factor of two faster and at least as accurate as the current state-of-the-art.
- We demonstrate the methods ability to generalize to more than five points and the increase in precision.
- We provide a novel type of dataset consisting of real image sequences and high accuracy ground truth camera poses (throw the authors homepage).

The remainder of the paper is structured as follows. In the subsequent section, we formalize the relative pose estimation problem and introduce our novel algorithm. In section three we describe the evaluation and in section four we present and analyze the results.

## 2. Methods

### 2.1. Problem formulation

A 3D world point $\mathbf{x}_i^w$ is projected onto two images, resulting in two corresponding image points $\mathbf{x}_i$ and $\mathbf{x}'_i$. The projections are defined by

$$\mathbf{x}_i \sim \mathbf{P}_1\mathbf{x}_i^w \ , \qquad \mathbf{P}_1 = \mathbf{K}_1[\mathbf{I}|\mathbf{0}] \qquad (1)$$

$$\mathbf{x}'_i \sim \mathbf{P}_2\mathbf{x}_i^w \ , \qquad \mathbf{P}_2 = \mathbf{K}_2[\mathbf{R}_{12}|\mathbf{t}_{12}] \ , \qquad (2)$$

where $\sim$ denotes equality up to scale and the first camera defines the coordinate system.

Let $\mathbf{R}_{12}$ and $\mathbf{t}_{12}$ denote the rotation and the translation between camera 1 and 2. The cameras are related as, [2],

$$\mathbf{x}'^T\mathbf{K}_2^{-1^T}\mathbf{E}\mathbf{K}_1^{-1}\mathbf{x} = 0 \ , \qquad (3)$$

where $\mathbf{E} = \mathbf{R}_{12}[\mathbf{t}_{12}]_\times$ is the essential matrix, which contains the relative rotation and translation between the views. The translation $[t_x \ t_y \ t_z]$ is represented as a skew symmetric cross product matrix

$$[\mathbf{t}_{12}]_\times = \begin{bmatrix} 0 & -t_z & t_y \\ t_z & 0 & -t_x \\ -t_y & t_x & 0 \end{bmatrix} \ . \qquad (4)$$

If the internal camera parameters $\mathbf{K}_1, \mathbf{K}_2$ are known, we can assume that the image points are rectified, i.e. , they are multiplied beforehand by $\mathbf{K}_1^{-1}$ and $\mathbf{K}_2^{-1}$, respectively. In that case, (3) becomes

$$\mathbf{x}'^T\mathbf{E}\mathbf{x} = 0 \ . \qquad (5)$$

To simplify notation, we omit the subindex $\cdot_{12}$ for the rotation $\mathbf{R}$ and the translation $\mathbf{t}$ in what follows.

### 2.2. Essential matrix estimation problem

Equation 5 allows us to verify whether the image points and the essential matrix are consistent, but in practice we need to estimate $\mathbf{E}$ from sets of points. Usually, these sets of points contain outliers and to obtain a robust estimate, one usually uses RANSAC. Within the RANSAC loop, it is preferable to use as few points as possible, [3]. The minimum number of points is determined by the degrees of freedom in $\mathbf{E} = \mathbf{R}[\mathbf{t}]_\times$ as (5) gives us one equation per correspondence. Hence we need five points.

The degrees of freedom also determine the dimensionality of minimal parameterizations. For reasons of computational efficiency, we chose a parametrization with five angles $\mathbf{w} = [\alpha \ \ \beta \ \ \gamma \ \ \theta \ \ \phi]$, further details are given in Appendix A.

Given five points, we aim at computing $\mathbf{E}(\mathbf{w})$ using an *iterative solver of non-linear systems of equations* instead of a closed form solution. For this purpose, we need to have geometric error rather than a purely algebraic one. One possibility to define a geometric error is in terms of the distance between a point in one view $\mathbf{x}_i$ and the epipolar line $\mathbf{l}_i$ from the corresponding point $\mathbf{x}'_i$ in the other view, [5]:

$$r_i(\mathbf{w}) = \mathbf{l}_i(\mathbf{w})\mathbf{x}_i \ , \qquad (6)$$

where

$$\mathbf{l}_i(\mathbf{w}) = \frac{\mathbf{x}'^T_i\mathbf{E}(\mathbf{w})}{||\mathbf{x}'^T_i\mathbf{e}_1(\mathbf{w}) \ \ \mathbf{x}'^T_i\mathbf{e}_2(\mathbf{w})||_2} \qquad (7)$$

and $\mathbf{e}_j(\mathbf{w})$ are the column vectors of $\mathbf{E}(\mathbf{w})$. Note that the symmetric variant of (6) is the Sampson distance, [13].

Hence, we want to solve

$$\mathbf{r}(\mathbf{w}^*) = \begin{bmatrix} r_1(\mathbf{w}^*) \\ \vdots \\ r_5(\mathbf{w}^*) \end{bmatrix} = \mathbf{0} \qquad (8)$$

and $\mathbf{w}^*$ is a global minimizer of

$$\sigma(\mathbf{w}) = \frac{1}{2}||\mathbf{r}(\mathbf{w})||_2^2 \ . \qquad (9)$$

### 2.3. Iterative solution using Dog Leg

Equation 9 can be categorized as a non-linear least squares problem, and solved by e.g. the Gauss-Newton method, [14]. We have instead chosen to apply an alternative method that combines Newton-Raphson (NR) and steepest descent (SD), called *Dog Leg* (DL), because "The Dog Leg method is presently considered as the best method for solving systems of non-linear equation." [8]

The combination of NR and SD is determined by means of the radius $\Delta$ of a *trust region*. Within this trust region, we assume that we can model $\mathbf{r}(\mathbf{w})$ locally using a linear model $\boldsymbol{\ell}$:

$$\mathbf{r}(\mathbf{w}+\mathbf{h}) \approx \boldsymbol{\ell}(\mathbf{h}) \triangleq \mathbf{r}(\mathbf{w}) + \mathbf{J}(\mathbf{w})\mathbf{h} \ , \qquad (10)$$

where $\mathbf{J} = [\frac{\partial \mathbf{r}}{\partial w_1} \dots \frac{\partial \mathbf{r}}{\partial w_5}]$ is the Jacobian.

The Newton-Raphson update $\mathbf{h}_{\mathrm{nr}}$ is obtained by solving $\mathbf{J}(\mathbf{w})\mathbf{h} = -\mathbf{r}(\mathbf{w})$ e.g. by Gauss elimination with pivoting. If the update is within the trust region ($\|\mathbf{h}_{\mathrm{nr}}\|_2 \leq \Delta$), it is used to compute a new potential parameter vector

$$\mathbf{w}_{\mathrm{new}} = \mathbf{w} + \mathbf{h}_{\mathrm{nr}} \ . \tag{11}$$

Otherwise, compute the gradient $\mathbf{g} = \mathbf{J}^T\mathbf{r}$ and the steepest descent step $\mathbf{h}_{\mathrm{sd}} = -\alpha\mathbf{g}$ (for the computation of the step length $\alpha$, see Appendix B). If the SD step leads outside the trust region ($\alpha\|\mathbf{g}\|_2 \geq \Delta$), a step in the direction of steepest descent with length $\Delta$ is applied

$$\mathbf{w}_{\mathrm{new}} = \mathbf{w} - (\Delta/\|\mathbf{g}\|_2)\mathbf{g} \ . \tag{12}$$

If the SD step is shorter than $\Delta$, $\beta$ times $\mathbf{h}_{\mathrm{nr}} + \alpha\mathbf{g}$ is added to produce a vector of length $\Delta$:

$$\mathbf{w}_{\mathrm{new}} = \mathbf{w} - \alpha\mathbf{g} + (\mathbf{h}_{\mathrm{nr}} + \alpha\mathbf{g})\beta \ . \tag{13}$$

In all three cases, the new parameter vector is only used in the next iteration ($\mathbf{w} = \mathbf{w}_{\mathrm{new}}$) if the gain factor $\rho$ is positive (for the computation of $\rho$ see Appendix B). Depending on the gain factor, the region of trust is growing or shrinking. The iterations are stopped, if $\|\mathbf{g}\|_\infty$, $\|\mathbf{r}\|_\infty$, $\|\mathbf{h}\|_2$, or $\Delta$ is below a threshold or if a maximum number of iterations is reached.

## 2.4. Initialization

As with any iterative method, the DL method needs a starting point $\mathbf{w}_0$. In the case of pose estimation for video frames its often possible to quite accurately guess the position of the next frame, because of physical limitations of motion (visual odometry). A good prediction is that the camera motion is constant between frames. To create a wider baseline for the pose estimation it is common not to take subsequent frames but frames further apart. In this case, the motion until the second but latest frame is already known and using the prediction described above, a good starting point for the pose estimation is obtained. This is meant when referring to **DLinit** in the Result section.

In certain cases, the initial position might be completely unknown, e.g. photo sharing sites. The main focus of this work is however towards real-time video sequences implying that we have some knowledge of the previous pose. The case without prior knowledge is still of occasional interest i.e. in an initial state or when one loses track resulting in a completely unknown pose. In these cases we obtain convergence by first initiating the solver with a standard parameter vector $\mathbf{w} = 0$ (corresponding to a forward motion) for a number of RANSAC iterations. After these first iterations, we initialize the solver with the currently best model (largest RANSAC consensus set). This is referred to as **DLconst** in section 4. The major advantage of DLinit compared to DLconst is that the individual RANSAC iterations are data independent and can easily be implemented on a parallel computer, e.g. GPU.

There are some further parameters in the DL algorithm that have to be chosen. $\Delta_0$ is the start value for $\Delta$, which is set to 1 in our implementation. An upper limit of the number of iterations is defined to avoid a waste of computational power in cases of that do not converge. In these cases, the resulting pose estimate is usually inaccurate and will be neglected anyway. We use an upper limit 5 for the number of iterations for DLinit. For DLconst we use a two step strategy: for the first 100 iterations the limit set to 8 and then it is lowered to 6. The thresholds for $\|\mathbf{g}\|_\infty$, $\|\mathbf{r}\|_\infty$, $\|\mathbf{h}\|_2$, and $\Delta$ are set to $10^{-9}$, $10^{-9}$, $10^{-10}$, and $10^{-10}$, respectively.

## 2.5. The processing pipeline

In order to apply the algorithm from the previous section to image sequences, several steps have to be performed before. First of all, we need to calibrate the camera which is done with Zhang's method, [19]. After this, distinct image regions are extracted using an interest point detector, [15].

We use the KLT-tracker, [15], to obtain point correspondences over time. The KLT-tracker is basically a least-squares matching of rectangular patches that achieves subpixel accuracy by gradient search. We use a translation-only model between neighboring frames. Tracking between neighboring frames instead of across larger temporal windows improves the stability, especially with respect to scale variations that are present in forward motion. When tracking frame-by-frame, the changes in viewing angle and scale are sufficiently small for tracking to work well. In order to improve stability of tracked points, the track-retrack (cross-check) has been applied in certain cases, [6]. All these steps were performed using Willow Garage's OpenCV implementations.

# 3. Evaluation setup

We have evaluated our method on three types of data, consisting of synthetic images, real images and the Notre Dame dataset [16]. Pose estimation methods are commonly tested on synthetic data with added artificial noise, [9, 12, 18, 17]. Since real data might differ significantly from synthetic test data, we have also evaluated our method on a dataset with real images.

The error metric used for our datasets are the difference in angle of the translation direction between the ground truth and the pose estimate. The direction is used because of the scale associated with the relative pose estimation is ambiguous. The Notre Dame data set is used to evaluate how well it generalizes to N-points and in order to compare it with the very accurate and fast method of Hartley et al. [4] we use an error metric based on differences in rotation.

In general, pose estimation methods behave differently on sideways motion and on forwards motion, [12, 18, 9]. In order to analyze both cases separately, we have divided the datasets into two parts, one for forward motion and one for sideways motion.

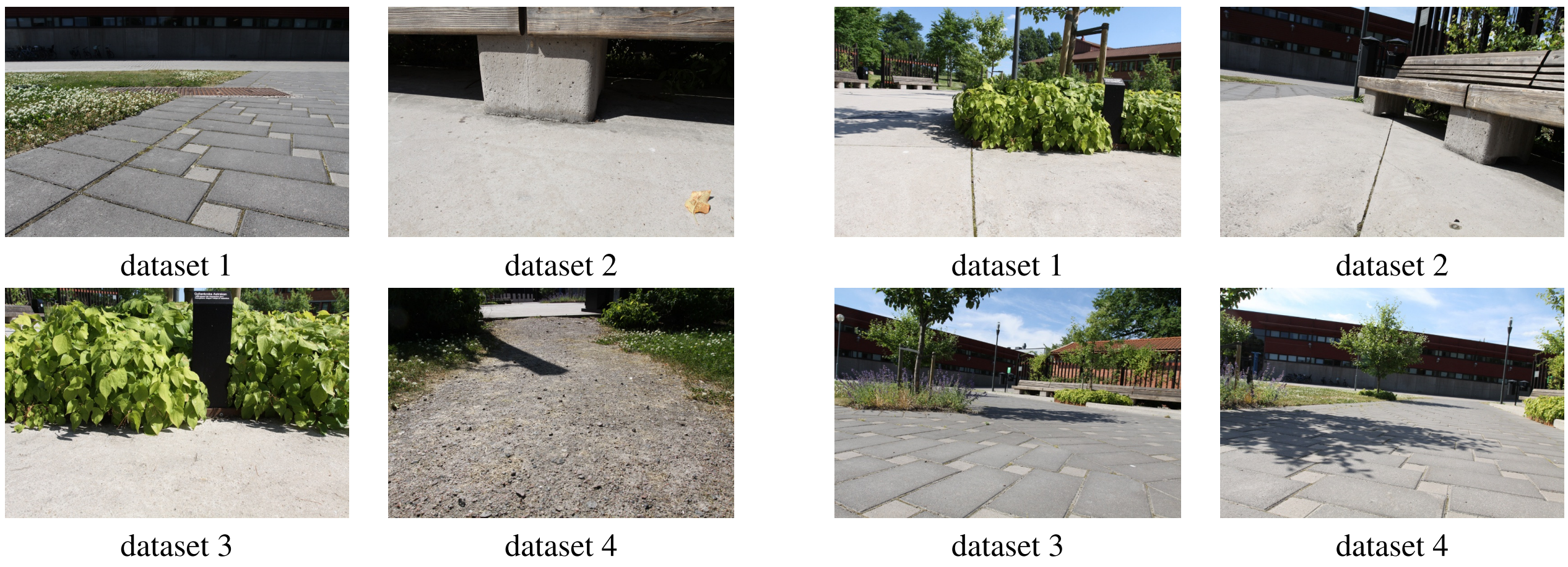

Figure 1. Forward motion, real data.

Figure 2. Sideways motion, real data.

### 3.1. Synthetic data

We have tested the algorithm on a synthetic dataset, which consists of 1000 randomly positioned 3D world points viewed by a camera moving both sideways and forwards. The world points were projected into the cameras and Gaussian noise (zero mean) was added to the projected point positions. The noise variance is varied between 0 and 1, where standard deviation 1 is equivalent to one pixel in an image with the size 1024x1024. The horizontal view-angle is set to 60 °. This follows the methodology used in [9, 18].

### 3.2. Real image sequences

The second evaluation dataset is constructed from real images by applying a methodology motivated by [1]. For capturing the images, a 15 megapixel Canon 50D DLSR camera has been used, which is capable to capture 6.5 images per second. The relatively low frame rate was compensated by moving the camera approximately one fourth of the wanted speed. By mounting the camera on a stable wagon and pushing it forward we could get a smooth camera trajectory for the sequence. This allowed us to get approximate ground truth for the sequence by estimating the pose for the high resolution images. A large quantity of points has been tracked (∼10K) and the pose has been estimated by using the five point method, [17] within a 10000 RANSAC iterations. Examples for the acquired images are shown in Figure 1 (forward motion) and Figure 2 (sideways motion). The dataset images for evaluation are obtained by down-sampling the images to 1 megapixel.

### 3.3. The Notre Dame dataset

The full Notre Dame data set consists of 715 images of 127431 3D points. The number of image pairs with more than 30 point correspondences are 32051. The dataset comes with already reconstructed cameras and points which has been refined with bundle adjustment. As in [4] we have chosen to use the bundle adjusted cameras as ground truth.

## 4. Results

The results are presented in Figures. 3-4 as box plots over 500 individual pose estimates. The boxes cover the second and third quantile and the dots are considered outliers. Each pose estimate is produced by running a 500 iterations RANSAC loop.

All methods are run on the same set of five points, which is randomly selected before each RANSAC iteration. Therefore, in the ideal case, all methods should produce the same result for a fixed number of RANSAC iterations. However, small deviations between the methods occur due to sporadic failures. Failures in Nistér's method are presumably caused by the preemptive scoring scheme, where all solutions are scored using a sixth point and the solution with the best score is chosen, [10]. Given that the sixth point is an outlier, a wrong solution is likely to be chosen and a potentially better solution and larger consensus set is potentially disregarded, but with respect to the whole RANSAC loop this is still the most efficient solution. The proposed method occasionally misses the correct solution due to slow convergence or convergence to another solution with the same consequences regarding the consensus set. The failure cases/deviations between the methods can be observed more frequently in hard cases such as the forward case with higher amounts of noise.

## 4.1. Synthetic data

Figure 3 shows the results on synthetic data. We have evaluated all methods on 4 different noise levels with standard deviation 0.25, 0.50, 0.75, and 1 pixel.

The size of the consensus set depends on the epipolar distance threshold that is used when evaluating the solutions. In order to get an appropriate fraction of points as inliers, we need to adjust the threshold according to the noise level. Without going into further details, the threshold has been scaled linearly with the noise. The noise level can be estimated using e.g. the forward-backward tracking error.

According to the box plots, the proposed method performs nearly identical to the method of Nistér for all datasets if ti is initialized with the previous camera position (DLinit).

One small performance increase can be observed for the forward case with large noise and the case of DLconst. This is the hardest case to solve and a high loss of solutions during the RANSAC iteration is expected. The increase of performance of DLconst is probably caused by selecting the currently best solution as initialization for the next RANSAC iteration. However, the improvement is data dependent and not statistical significant.

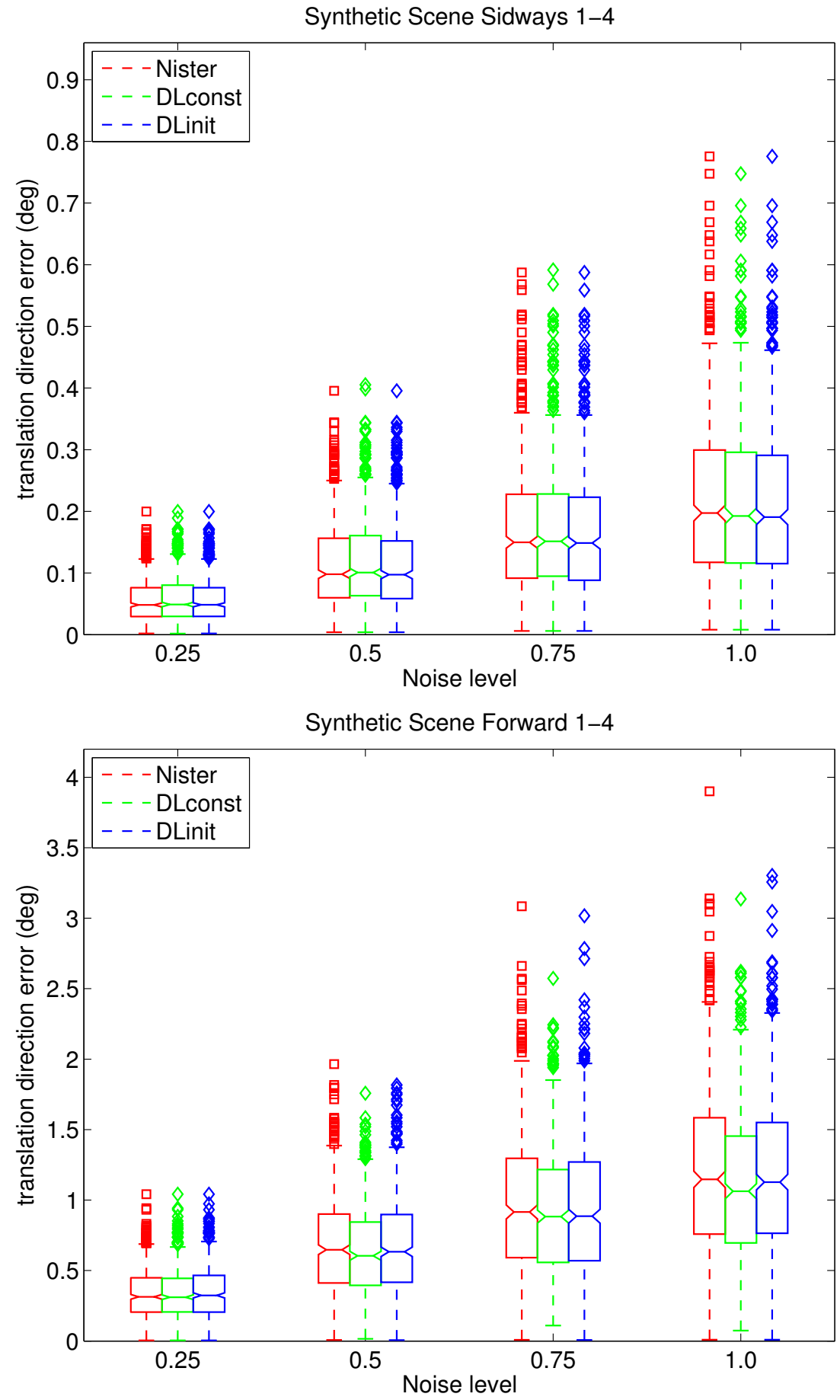

Figure 3. Synthetic data

## 4.2. Real image sequences

In contrast to the synthetic data, the noise level in the real data is approximately constant and therefore the same threshold for the epipolar distance is used all cases.

Overall, the accuracy is nearly identical for all methods. Despite the quite similar performance for the same number of RANSAC iterations, the DL method is superior in the way that it is faster than Nistér's method, see next section.

Figure 5 shows the difference in precision between Nistér's and the proposed method. For each RANSAC range of 100 iteration, it shows the error of the currently best solution from Nistér's algorithm minus the error of the currently best solution from our method. Bars in the negative range indicate that Nistér's result is more accurate and bars in the positive range indicate that our results are better. As can be seen, in all cases the median error is near zero. Note that the error difference is in the order of $10^{-6}$ to $10^{-8}$, which is several orders of magnitude smaller than the error in the pose estimate (about 0.4 degree). Practically speaking, the methods perform equally well in terms of precision for the evaluated datasets. One exception is observed for the forward motion and DLconst after 100 iterations, where one quantile lies significantly below 0 for datasets 3 and 4 (not the median though). This is presumably caused by a failed initialization using the DLconst strategy.

## 4.3. Speed

The faster the method, the more RANSAC iterations can be computed in the same time. The more RANSAC iterations are computed, the better precision can be achieved, [5]. This is why the method of [9] is still considered to be state-of-the-art, even if more accurate methods like [17] have been proposed later.

Our method is implemented in plain C/C++ code. It is all compiled and tested on a desktop PC with an Intel 2.66Ghz (W5320) CPU.

The proposed method is compared with 2 implementations of Nistér's algorithm, one which is contained in VW34 (a library for real-time vision developed at Oxford's Active Vision Lab[1]). The second is from Richard Hartley and is to our knowledge the fastest available implementation [2].

To accurately compare our method with Nistér's implementation, we have compiled our code on a Pentium III desktop PC, with the same type of processor that is used in [9]. The performance timings are then compared with the ones found in [9].

The average speedups on our data sets are summarized

[1]http://www.robots.ox.ac.uk/ActiveVision/
[2]http://users.cecs.anu.edu.au/ hartley/

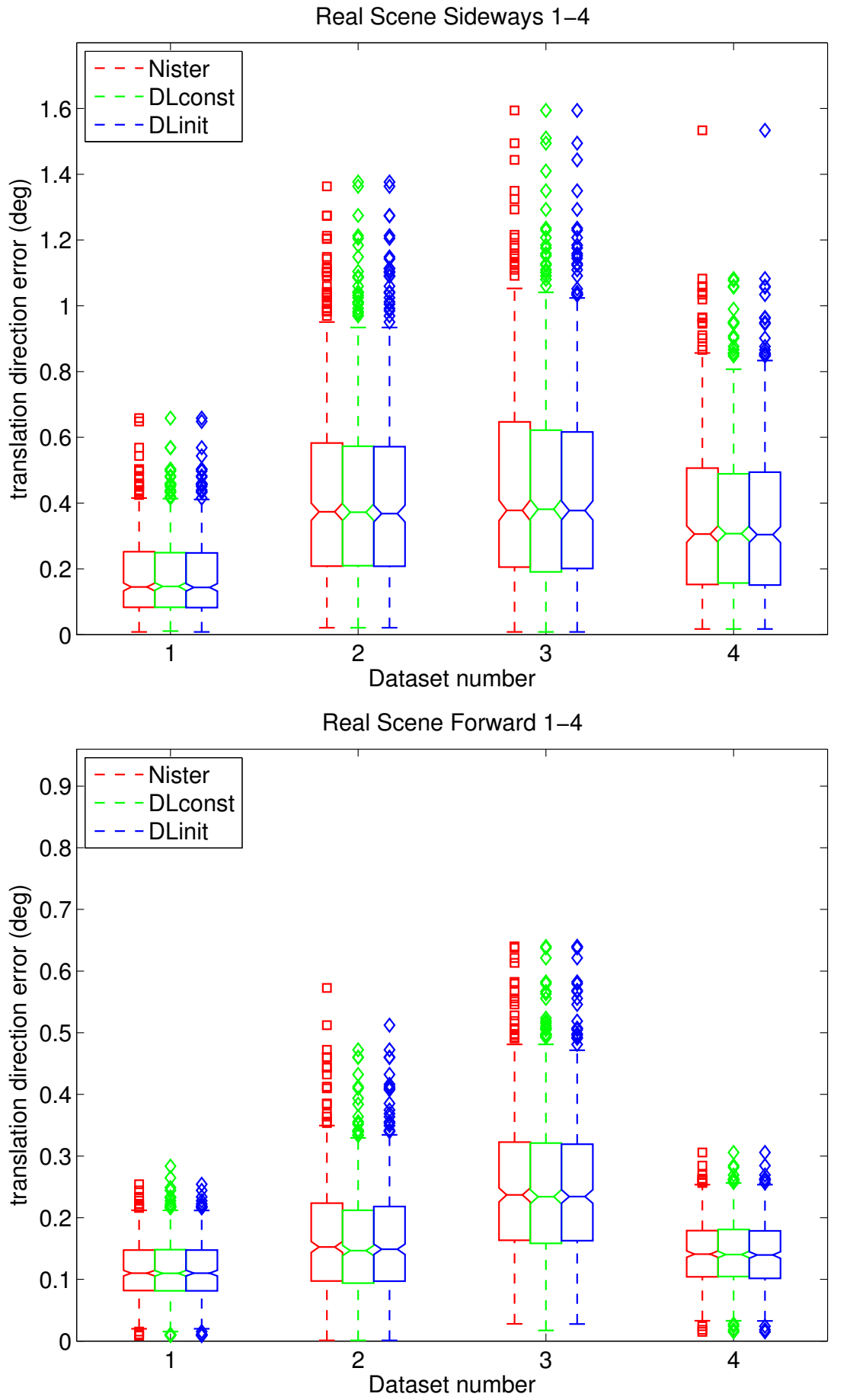

Figure 4. Real data

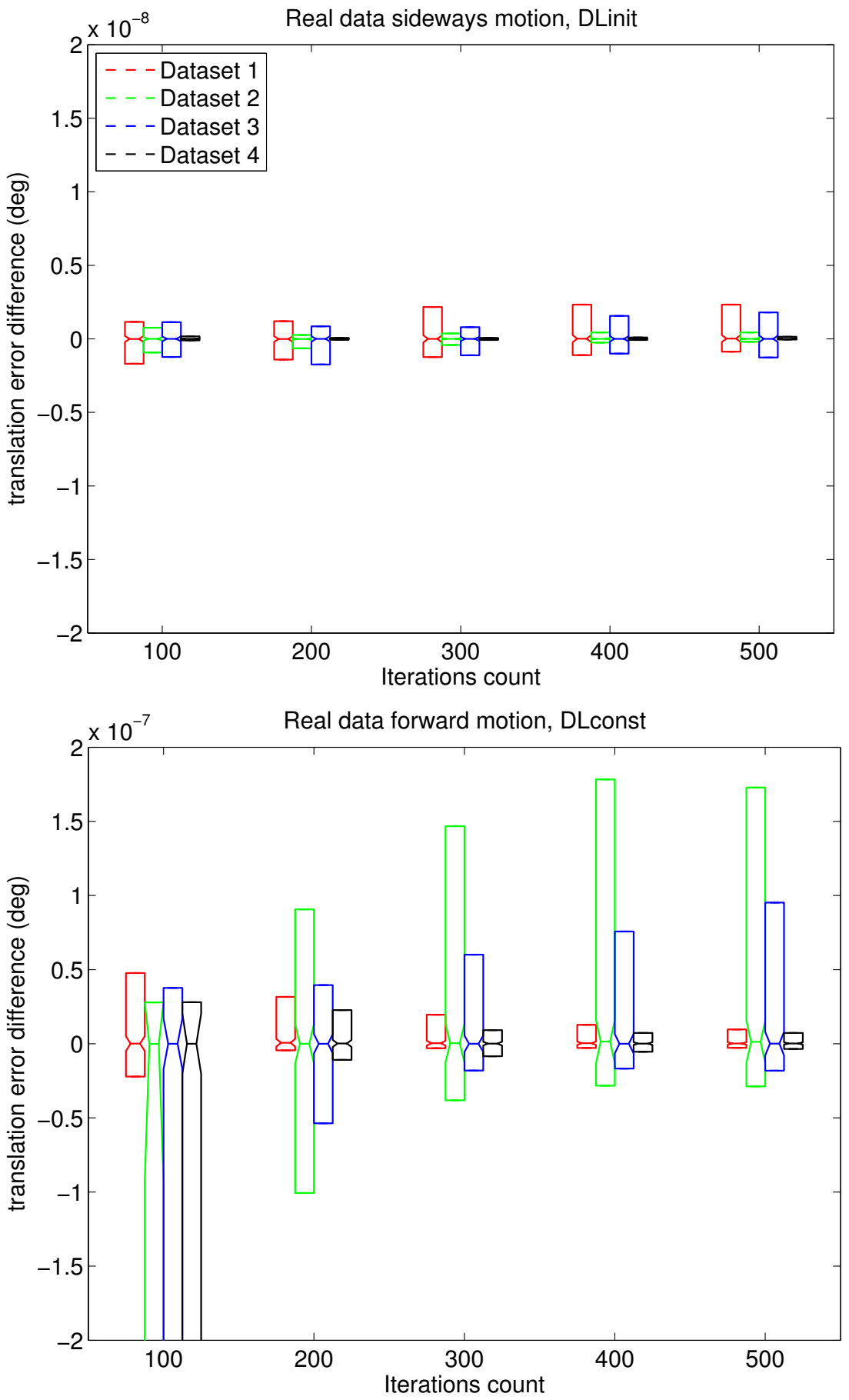

Figure 5. The bar plots shows the difference in precision of every new best solution found whit in the RANSAC loop. Here we show the best and the worst case for our method on real data.

| Method | i7 | PIII 550 |
|---|---|---|
| [9] | x | 121$\mu$s |
| Nistér(available) | 16,5$\mu$s | x |
| VW34 | 34.8$\mu$s | x |
| Ours (mean) | **7.0**$\mu$s | **72.1**$\mu$s |

Table 1. Timings for different methods and implementations. The time for our method is the mean over all tested datasets.

| Data type | time | Nistér(paper) | Nistér(imp) |
|---|---|---|---|
| Realdata DLconst | 7.5$\mu$s | 1.6x | 2.2x |
| Realdata DLinit | 6.5$\mu$s | 1.9x | 2.5x |
| Syntdata DLconst | 7.7$\mu$s | 1.5x | 2.1x |
| Syntdata DLinit | 7.1$\mu$s | 1.7x | 2.3x |

Table 2. Average timings for each of the real and synthetic datasets, using DLconst and DLinit. The Nistér(paper) column is the speed up compared with [9] on the PIII and Nistér(imp) is the implementation freely available from Hartley

in table 1. For a modern CPU and a currently available code we are on average 2-2.5 times faster than the closed form solution. Compared with Nistér's own implementation on older hardware, we achieve an average speedup between 1.5x and 1.9x.

### 4.4. N-point generalization

The approach in section 2.3 allows us to use more then five points to solve the relative pose, with very minor changes to the actual implementation. Instead of solving the square linear equation (LSQ) system $\mathbf{J}(\mathbf{w})\mathbf{h} = -\mathbf{r}(\mathbf{w})$ arising from (10), we need to solve the alternative LSQ system $\mathbf{J}(\mathbf{w})^T\mathbf{J}(\mathbf{w})\mathbf{h} = -\mathbf{J}(\mathbf{w})^T\mathbf{r}(\mathbf{w})$ (a.k.a. the normal equations). This and the fact that the residual is defined as an reprojection error (line to point distance in image), renders our method an interesting alternative for overdetermined relative pose problems.

The current state-of-the-art method for using more than five points for determining rotations is due to Hartley et al.

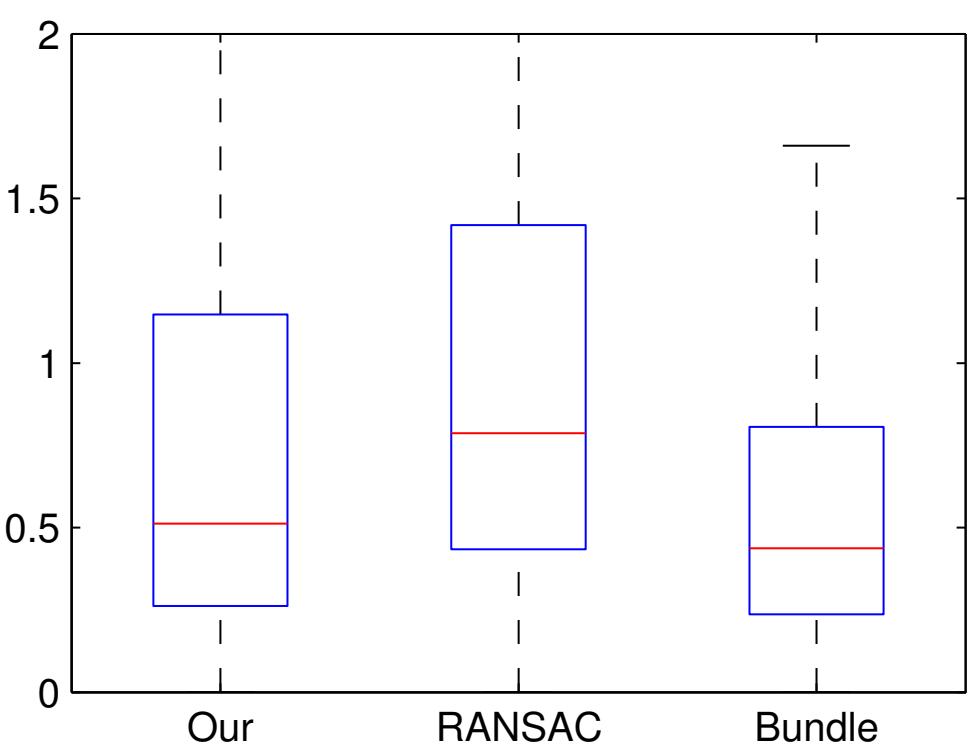

Figure 6. A box plot over the precision for three different method, our, standard five point solver within a RANSAC framework and 2-view bundle adjustment. The line in the box marks the median and the top and bottom of the boxes are the 25th and 75th percentiles.

[4]. The method consists of an L1-rotation averaging of several five-point pose estimates and the authors show how this can be used to fast and accurately estimate relative rotations for the Notre Dame dataset [16] under 3 minutes.

We have compared the generalization to N-points of our method to the result in [4]. Our approach consists of two steps, first a small set of RANSAC iterations (20 with DLinit) are run with our method using five points. Then the final pose/rotation is estimated using N points, where N is the number of correspondences in the current image pair. As in [4] we only use image pairs with more than 30 correspondences.

In our experiment, we use the full 715 images Notre Dame data set to compare our N-point approach, [9] within a RANAC framework and the 2-view bundle adjustment. Table 4.4 compares our result to the one in [4]. The precision is shown as the increase in angular error compared to using a 2-view bundle adjustment approach. Both methods has high precision and come within 20% of the angular error of the much heavier bundle approach. Hovever as seen in table 4.4 our method is by far the fastest (7.5x faster than current state-of-the-art) while also retaining the best estimating (closest to the 2-view bundle adjustment).

| Data type | time | precision |
|---|---|---|
| Our | 22s | 15.9% |
| Hartley | 168s | 19.5% |
| 2-view BA | 11932s | 0.0% |

Table 3. Trimmings and precision for Hartley and 2-view BA are from [4]. [4] does not use the same CPU model as here but has the same clock frequency (2.6Ghz). The relative precision error is defined as the increase in angular error as compared to 2-view bundle adjustment (in L1-norm).

## 5. Conclusion

We have presented a new iterative method for solving the relative pose problem for five points. The method is based on Powell's DogLeg algorithm and achieves at least the same accuracy as state-of-the-art methods on both synthetic and real data. The proposed algorithm is approximately twice as fast as Nistér's method, depending on hardware and implementation, which can be exploited within the RANSAC algorithm to increase the number of iterations and thus to further enhance accuracy. In addition, the proposed method has approximately half the number of code lines compared to the implementation from [4] which simplifies implementation and maintenance of code.

In contrast to a closed for solver such as, [9] the presented method can solve the pose for more points than five with very minor changes to the current implementation. We have showed that this generalization is highly competitive with the current state-of-the-art [4], having similar accuracy while being considerably faster.

**Acknowledgements** The authors gratefully acknowledge funding from the Swedish Foundation for Strategic Research for the project VPS, through grant IIS11-0081. Further fundings have been received through the ECs 7th Framework Programme (FP7/2007-2013), grant agreement 247947 (GARNICS), and from the project Extended Target Tracking founded by Swedish Research Council.

## Appendix A: parametrization of the essential matrix and computation of the Jacobian

The essential matrix can be factorized as $\mathbf{E} = \mathbf{R}[\mathbf{t}]_\times$. The rotation $\mathbf{R}$ is parametrized using angles, $\mathbf{R}(\alpha,\beta,\gamma) = \mathbf{R}_x(\alpha)\mathbf{R}_y(\beta)\mathbf{R}_z(\gamma)$, where $\mathbf{R}_x(\alpha)$ is the rotation matrix around the x-axis and equally $\mathbf{R}_y$, $\mathbf{R}_z$ for the y- and z-axis.

The motivation for choosing this particular representation instead of other options, e.g. quaternions or Rodrigues parameters, is computational speed. Using angles reduces the number of arithmetic operations for computing the derivatives by at least a factor of two.

The translation vector $\mathbf{t}$ of the relative pose is only defined up to scale. In order to have a minimal representation, the translation vector is fixed to unit length by representing it in spherical coordinates $\mathbf{t}(\theta,\phi) = [\sin(\theta)\cos(\phi) \quad \sin(\theta)\sin(\phi) \quad \cos(\theta)]^T$.

Parametrizing the essential matrix with $\mathbf{w}$ gives $\mathbf{E}(\mathbf{w}) = \mathbf{R}(\alpha,\beta,\gamma)[\mathbf{t}(\theta,\phi)]_\times$. We further reformulate (5):

$$\mathbf{x}'^T\mathbf{E}\mathbf{x} = \tilde{\mathbf{E}}\mathbf{q}, \quad \text{where } \mathbf{x} = [x\ y\ 1]^T \text{ and}$$

$$\mathbf{q} = \begin{bmatrix} x'x & y'x & x & x'y & y'y & y & x' & y' & 1 \end{bmatrix}^T$$

$$\tilde{\mathbf{E}} = \begin{bmatrix} e_{11} & e_{12} & e_{13} & e_{21} & e_{22} & e_{23} & e_{31} & e_{32} & e_{33} \end{bmatrix}$$

The Jacobian now reads

$$\frac{\partial r_i}{\partial w_j} = \frac{\partial}{\partial w_j}(\frac{1}{\sqrt{s}}\tilde{\mathbf{E}}\mathbf{q}_i) \qquad i,j \in \{1\dots5\} \tag{14}$$

where

$$s = (x'e_{11} + y'e_{21} + e_{31})^2 + (x'e_{12} + y'e_{22} + e_{32})^2.$$

The product rule gives

$$\frac{\partial r_i}{\partial w_j} = \frac{1}{\sqrt{s}}(\frac{\partial \tilde{\mathbf{E}}}{\partial w_j} - \frac{1}{2s}\frac{\partial s}{\partial w_j}\tilde{\mathbf{E}})\mathbf{q}_i \ , \ \text{where} \tag{15}$$

$$\frac{\partial s}{\partial w_j} = 2(x'e_{11} + y'e_{21} + e_{31})(x'\frac{\partial e_{11}}{\partial w_j} + y'\frac{\partial e_{21}}{\partial w_j} + \frac{\partial e_{31}}{\partial w_j}) + 2(x'e_{12} + y'e_{22} + e_{32})(x'\frac{\partial e_{12}}{\partial w_j} + y'\frac{\partial e_{22}}{\partial w_j} + \frac{\partial e_{32}}{\partial w_j}).$$

## Appendix B: algorithmic details of DL

The optimal step length $\alpha$ for the gradient descent is determined as, [8]

$$\alpha = \frac{\|\mathbf{g}\|_2^2}{\|\mathbf{J}(\mathbf{w})\mathbf{g}\|_2^2} \ . \tag{16}$$

The gain factor is determined by comparing the factual gain with the gain according to the model

$$L(\mathbf{h}) = \frac{1}{2}\|\mathbf{r}(\mathbf{w}) + \mathbf{J}(\mathbf{w})\mathbf{h}\|_2^2 \ , \tag{17}$$

and

$$\rho = (\sigma(\mathbf{w}) - \sigma(\mathbf{w}_{\text{new}}))/(L(\mathbf{0}) - L(\mathbf{h}_{\text{dl}})). \tag{18}$$

If $\rho > 0.75$, the trust region radius $\Delta$ is growing to $\max\{\Delta, 3\|\mathbf{h}_{\text{dl}}\|_2\}$. If $\rho < 0.25$, the trust region is shrinking to $\Delta/2$.

The matrices handled here have so small dimensions that it is more effective to implement the linear algebra operation yourself (using standard implementation scheme with fixed looping) then to use optimized linear algebra packages, e.g. Lapack, due to their large overhead.

A further performance improvement is achieved by reducing the number of trigonometric instructions that are used. When updating $\mathbf{w}$ with $\mathbf{h}$, many of the update steps are very small and trigonometric updates can be computed by using the sum of angles formula for $\sin()$ and $\cos()$ in combination with a Taylor expansion of $\sin(w_j)$ and $\cos(w_j)$.